# AI assurance using causal inference: application to public policy

Andrei Svetovidov[a], Abdul Rahman[a], Feras A. Batarseh[a,b]
[a] *Commonwealth Cyber Initiative, Virginia Tech, Arlington VA, USA*
[b] *Bradley Department of Electrical and Computer Engineering, Arlington VA, USA*

## Highlights

In this chapter, the following topics are covered:
- Fundamentals of the causal inference theory
- Review of the concept of AI assurance and its connection to causality
- Assurance-focused causal experiment on the internet speed dataset
- Methods of graph-based data representation and analysis

# Graphical abstract

**AI system**

→ **Causal Inference**
- Foundations
- Average Treatment Effect
- Literature review graph

→ **AI Assurance**
- Systems fairness and explainability

**AI Assurance Using Causal Inference**
- Observed causal relationships (internet speed dataset)

```
Refute: Use a Placebo Treatment
Estimated effect:828406.606918239
New effect:2876.0043238992657
p value:0.4749805528770472
```

| State   | Year | Funds | Population_density | Treatment |
|---------|------|-------|--------------------|-----------|
| Alabama | 2013 | 24296 | 95.370708          | FALSE     |
| Alabama | 2014 | 24577 | 95.602082          | TRUE      |
| Alabama | 2015 | 24546 | 95.810354          | FALSE     |
| Alabama | 2016 | 26474 | 96.031065          | TRUE      |
| Alaska  | 2013 | 11288 | 1.291649           | FALSE     |
| Alaska  | 2014 | 11397 | 1.290274           | FALSE     |
| Alaska  | 2015 | 13767 | 1.292403           | FALSE     |
| Alaska  | 2016 | 10602 | 1.299339           | FALSE     |
| Arizona | 2013 | 27668 | 58.39005           | FALSE     |
| Arizona | 2014 | 28905 | 59.249681          | FALSE     |
| Arizona | 2015 | 31182 | 60.123521          | FALSE     |

→ **Data as a network**
- Graph theory introduction
- Examples of networks

# AI assurance using causal inference: application to public policy

Andrei Svetovidov[a], Abdul Rahman[a], Feras A. Batarseh[a,b]
[a] *Commonwealth Cyber Initiative, Virginia Tech, Arlington VA, USA*
[b] *Bradley Department of Electrical and Computer Engineering, Arlington VA, USA*

*Abstract*
Developing and implementing AI-based solutions help state and federal government agencies, research institutions, and commercial companies enhance decision-making processes, automate chain operations, and reduce the consumption of natural and human resources. At the same time, most AI approaches used in practice can only be represented as "black boxes" and suffer from the lack of transparency. This can eventually lead to unexpected outcomes and undermine trust in such systems. Therefore, it is crucial not only to develop effective and robust AI systems, but to make sure their internal processes are explainable and fair. Our goal in this chapter is to introduce the topic of designing assurance methods for AI systems with high-impact decisions using the example of the technology sector of the US economy. We explain how these fields would benefit from revealing cause-effect relationships between key metrics in the dataset by providing the causal experiment on technology economics dataset. Several causal inference approaches and AI assurance techniques are reviewed and the transformation of the data into a graph-structured dataset is demonstrated.

## 1. Introduction and Motivation

Artificial Intelligence (AI) has experienced a tremendous growth trend during the last decade due to availability of multivariate large-scale datasets and advancements in high-performance computations with multi-core GPUs. AI methods, such as classic machine learning algorithms, deep neural networks, and reinforcement learning, demonstrated impressive results in solving prediction and classification tasks in many domains, including transportation, healthcare, and finance (Boire, 2018). However, utilization of such methods is heavily underexplored in policy making. The institutions and agencies participating in legislature activities would clearly benefit from using cutting-edge AI models to make the lawmaking process more effective and be able to better address goals of government and state level officials (Zuiderwijk, Chen, and Salem, 2021).

At the same time, the lawmaking process itself is complex in nature, involves multiple steps, and can also vary from state to state or even at higher granularity. Moreover, the decisions taken by officials in the forms of issued policies determine the path of development of the country and its inhabitants. We have seen many examples of how passed laws influenced the lives of millions of people, such as the Coronavirus Act (Coronavirus Preparedness and Response Supplemental Appropriations Act, 2020). The law facilitated the production and distribution of COVID-19 tests and vaccines and allocated $22 billion in funding, which led to a faster vaccination across the United States and therefore building collective immunity against the disease. Since AI-based legislation systems are to be built on such sensitive and influential information, it is very important to assure their trustworthiness, transparency, and fairness towards the residents. If it is possible to explain how AI methods work and why they generate such results, we could maintain public trust in them and would eventually fully integrate such systems into policy making cycles.

One of the ways to leverage AI assurance in this scenario is to consider causal inference methods applied to the metrics of interest. For instance, can we infer that a cause-effect relationship exists between the proposed COVID-19 vaccine distribution law and the number of current positive cases? Another example is whether the regulations issued by the U.S. Department of Transportation lead to reduced driving times on target roads. Such dependencies between different factors are typically not captured by classic machine learning algorithms and more sophisticated methods are required. In addition, it would be helpful to also consider dependencies between input data vectors based on some other contextual information. For example, the laws approved and passed in separate states might influence each other to some degree if the states are geographically close to each other, situated in the same region of the U.S., or governed by the same political force. The aspects mentioned above should be carefully considered while creating an AI assurance model for policy making (Perry and Uuk, 2019).

In the next section of this chapter, we introduce the concept of causal inference and provide a detailed overview of some of the methods with an accompanying knowledge graph.

## 2. Causal Inference

Machine Learning (ML) and Deep Learning (DL) algorithms are heavily used as core elements of AI systems since they bring a power of detecting input/output data relationships and predicting the outcome for future inputs. However, a significant limitation of ML models is their inability to account for mutual correlation or association between inputs and outputs, leaving aside more complex dependencies. Thus, cause-effect relationships cannot be captured by a feed-forward neural network and applying this approach directly to some problems can lead to false inferences. Let us take one example: a user notices that their screen freezes every time they open Google Chrome on a personal computer. The user may assume that the problem is caused by launching the browser, whereas the underlying reason might be that there are too many active background OS processes which utilize most of available RAM, so that the system cannot easily handle such a "RAM-hungry" web browser as Chrome. If our neural network is trained on the results of user experience surveys to predict whether the system would freeze, it would produce wrong predictions for users whose RAM is not overloaded with running OS processes.

The example provided above clearly proves the famous statement: "*correlation does not imply causation*" (Aldrich, 1995). Indeed, there is an obvious correlation between crashing an OS and launching a web browser, although one does not cause another. In other settings, such as healthcare or national security, such incorrect inferences may lead to dramatic consequences involving safety and lives of people, as mentioned by Hamid (2016), so it is necessary to understand causal relationships before applying common AI solutions. In Section 2.1, we introduce the concepts of causality and causal inference and their mathematical foundations.

## 2.1. An Introduction to Causal Inference

Causal inference is a relatively new field of study within AI context but showed to be very promising in recent years. Although first inference engines were introduced as a main block of expert systems in 1970-1980s (Hayes-Roth, Waterman, and Lenat, 1983), they were not truly intelligent in a sense that they relied on predefined logical rules and were not capable of performing data-driven knowledge inference

Let us start with explaining the difference in terms used in this area. *Causality* is an existing relationship between the effect and what it was caused by (object, state, or process). *Causation* means the act of causing a particular event/state/process (Honderich, 1988). Although these two terms are often used interchangeably, *causation* can be viewed as a process of initiation of *causality.* For instance, the rain caused city dwellers to take out and open their umbrellas, which shows the relationship of causality between these events. The causation occurred right after the rain started making people use their umbrellas. *Causal Inference* is a field of study which attempts to reveal causal relationships between nodes via making causal assumptions (Pearl, 2009).

Causal inference methods were heavily involved in addressing challenges in healthcare (Moser, Puhan, and Zwahlen, 2020); therefore, they inherited some terminology from the latter. In addition to standard machine learning concepts of input data *X*, known as *covariate features of a patient,* and output data *Y*, known as

*outcome,* we also introduce *T - treatment -* the action taken on a patient (or, in medical terms, a treatment that was given to them). The relationships between *X*, *T*, and *Y* are usually presented as a Directed Acyclic Graph (DAG). In general, DAG represents complex causal structures, can be very cumbersome, and includes multiple nodes of each type, but for the sake of simplicity let us consider the following scenario: how a certain medication affects the patient's blood pressure level (Szolovits and Sontag, 2019). Here, *X* is the information about the patient known beforehand, *T* is a medication, which can be either *T0* (control treatment) or *T1* (actual medication), and *Y* is the patient's blood pressure after being treated. The corresponding causal graph is shown in Figure 1.

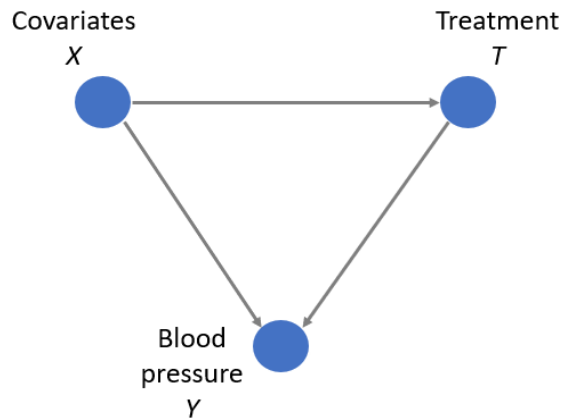

*Figure 1. Causal DAG example*

For each individual, we want to understand if the hypertension treatment actually works. For this we need to compare the effects of each treatment on a patient, which can be done via calculating Conditional Average Treatment Effect as per Neyman-Rubin Causal Model (Sekhon, 2008):

$$CATE(x_i) = E_{Y_1 \sim p(Y_1|x_i)}[x_i] - E_{Y_0 \sim p(Y_0|x_i)}[x_i] \quad (1),$$

where $E_{Y_1 \sim p(Y_1|x_i)}[x_i]$ and $E_{Y_0 \sim p(Y_0|x_i)}[x_i]$ is the expectation of the outcome had the individual $x_i$ had and had not been treated respectively. $x_i$ is a set of features of the patient.

The Average Treatment Effect for the entire population can be calculated as the expectation over CATE values for all instances:

$$ATE = E_{x \sim p(x)}[CATE(x_i)] \quad (2)$$

However, it is often impossible to measure the outcome of both treatments applied to the same individual due to many reasons, including safety and ethics, which is known as the fundamental problem of causal inference (Holland, 1986). Hospitals make educated decisions on what treatment to deliver to each patient, and providing several treatments simultaneously can not only ruin the reputation of the doctor or hospital and healthcare system in general, but lead to malicious effects on patient's health and violate certain regulations. Therefore, causal inference based on counterfactuals is only dealing with the data obtained from control and treatment groups (Szolovits and Sontag, 2019):

$$ATE = E_{x \sim p(x)}[\, E[Y_1 \mid x, T = 1] - E[Y_0 \mid x, T = 0]\,] \quad (3)$$

where $Y_1$ and $Y_0$ are the responses of patients being provided $T_1$ and $T_0$ respectively. As explained by Szolovits and Sontag (2019), this approach requires several strong assumptions to be held:

- Ignorability: there are no unobserved confounding variables. The mathematical form of this assumption shows that the outcome is independent of treatment given input data:

$$(Y_0, Y_1) \perp T \mid x \quad (4)$$

- Common support: there is always a stochasticity in treatment decisions:

$$p(T = t \mid X = x) > 0 \; \forall \, t, x \quad (5)$$

For example, we have a subpopulation of individuals with red hair. These should include patients from both control and treatment groups to satisfy this assumption.

- Stable Unit Treatment Value Assumption (SUTVA): the response (outcome) for a particular individual to a provided treatment is independent of treatments of other units.

Once the conditions mentioned above are satisfied, we can attempt to apply ML in a standard way to draw the relationships between inputs and outputs. It should be mentioned that these assumptions do not perfectly hold in the real world and inferring CATE values is still a probabilistic process. In Section 2.2, we overview state-of-the-art causal inference approaches that address these limitations.

2.2. Overview of Causal Inference Methods

In Section 2.1 we described the conditions that need to hold to apply ML methods for causal inference. One of the first approaches for estimating Average Treatment Effect (ATE) is called covariance adjustment (Szolovits and Sontag, 2019). In this method, a parametric model is being fitted on training data:

$$f(x, t) \approx E[Y_t \mid T = t, x] \quad (6),$$

where $f(x, t)$ is the function for approximation of the expectation $E[Y_t \mid T = t, x]$, and t equals 0 or 1 in binary setting.

Once the model is trained, we can calculate ATE estimation via finding the average difference between function values for each patient with treatment values 1 and 0 accordingly

$$\widehat{ATE} = \frac{1}{n} \sum_{i=1}^{n} \quad f(x_i, 1) - f(x_i, 0) \quad (7)$$

In the simplest case, there is a linear dependency between outcome and covariates/treatment described by the following equation:

$$Y_t(x) = \alpha x + \beta t + \gamma \quad (8),$$

where $\alpha, \beta, \gamma$ are trainable parameters.

Although this linear regression model is comparatively easy to train, it may oversimplify true causal relationships and eventually lead to incorrect assumptions. It is very important to ensure that the model has sufficient representative power to infer correct outcomes. Another reason for making incorrect assumptions - which is very common in practice - is violation of common support. In many domains and for many reasons it may not be feasible to draw data from the Randomized Control Trial type of experiment, where the patients are assigned treatment randomly. Therefore, there is an approach to estimate ATE while mitigating the selection bias, which is called propensity score matching (Szolovits and Sontag, 2019). It requires the following steps:

1. Estimate propensity scores for each individual, i.e. define the groups of similar patients in terms of provided treatment: $p(T = t \mid x)$
2. Match propensity scores: there are several methods, but the most popular one is based on nearest neighbor search.
3. Evaluate the matching technique and apply a different one if the results are unsatisfactory.
4. Calculate ATE for each stratum with the formula mentioned in Section 2.1 (group of individuals with close propensity scores) and take average or weighted average of them to estimate ATE for the entire population.

If the assumption of *ignorability* holds, i.e. data collection being independent of missing data, this method can balance initial bias in treatment assignment prior to estimating ATE.

The precursor of causal inference methods was the developed model called Bayesian network. It represented conditional dependencies of variables in a form of DAG. Even though Bayesian DAG cannot be used directly to identify causal effect, it gave rise to further developments in causality. One of relatively early causal inference methods - Bayesian Additive Regression Trees (BART) - uses decision tree algorithm as a building block, where the path is determined by conditions on X and T, and the value of Y is found at the end point of each path (Hill, 2011). Aggregate outcome from separate single trees is deemed a final result. This method predicted ATE more accurately than linear regression and propensity score matching but was very sensitive to the amount of training data. Künzel et al. (2019) proposed a more sophisticated method, in which the model consists of base learners: BART and causal forests. This approach provides a richer representation thanks to estimating an outcome for a treated individual using control-outcome estimator, and vice versa, which helps to account for imbalance between treatment groups.

Better results can be achieved with deep learning. In (Johansson, Shalit, and Sontag, 2016), researchers applied deep neural networks to counterfactual inference. Authors presented a modified version of this approach called TARNet in (Shalit, Johansson, and Sontag, 2019), where the network was augmented to the neural network with two parallel blocks to estimate the effect under treatment and control respectively. Also, the treatment assignment bias is adjusted by adding IPM (Integral Probability Metric) term, and the final objective function is a trade-off between treatment imbalance and accuracy. The technique proposed by Shalit et al. (2019) was utilized in other works, including (Schwab, Linhardt, and Karlen, 2018), where TARNet was extended to the algorithm with the ability to handle multiple different (non-binary) treatments and augment insufficient input data. Schwab et al. (2020) improved this work, where the joint neural network considers "dosage", or real-value treatment. The network structure of the dataset was considered in (Guo, Li, and Liu, 2020), where TARNet was combined with GNN to predict ATE. Yao et al. (2018) also took into account connections between instances (individuals) via implementation of local similarity information along with treatment distribution balancing and deep learning ATE estimator.

In multiple works, researchers extended ideas of counterfactual inference to more advanced cases. Hartford et al. (2017) introduced a two-stage DNN-based model which accounts for presence of hidden confounders via using instrumental variables. In (Lim, Alaa, and van der Schaar, 2018), researchers developed a framework for predicting an outcome of a chain of treatments. Kobrosly (2020) developed a Python package to estimate a dose-response curve with the Generalized Propensity Score and Targeted Maximum Likelihood Estimation. Louizos (2017) built a causal effect variational autoencoder to combine advancements of latent variables in machine learning and proxy variables utilization in causality. Rakesh et al. (2018) extended this work to also consider the pairwise spillover effect between covariates. Some of the works represent domain adaptations of causal inference methods, such as

(Bonner and Vasile, 2018), where recommendation policy optimization is done via increasing the desired outcome with ITE modeling.

The graphical representation of this literature overview is demonstrated in Figure 2 and Figure 3. All the works are into groups in accordance with their relation to causal inference:
- Independent and identically distributed (i.i.d.) data with binary treatment
- i.i.d. data with non-binary treatment (advanced cases)
- Domain-specific

The graph reveals a "parent-child" dependency of papers based on implementation and modification of ideas of some works in others. Edges of the connecting lines are also accompanied with short work description (in non-bold italic). For the sake of readability, the graph is broken down into two parts.

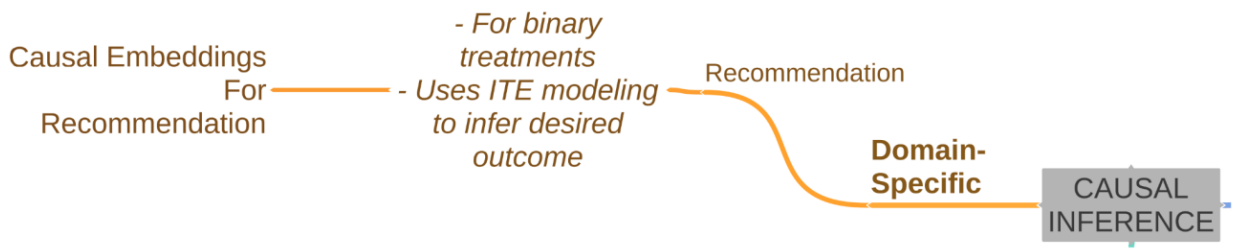

*Figure 2. Domain-specific part of literature review knowledge graph*

## 3. AI Assurance Using Causal Inference

In the previous section, we presented the fundamentals of causal inference and introduced a review of some recent works in the area. This section familiarizes the reader with the main ideas of AI Assurance (Section 3.1), and the authors attempt to connect these two areas of study by providing an experiment on tech-econ policy dataset (Section 3.2).

### 3.1. AI Assurance: goals and methods

As we mentioned earlier, assuring AI systems is becoming more necessary due to the growing demand in such systems, as well as increasing requirements to their transparency. For many critical applications, such as healthcare or military operations, there are several key assurance pillars that AI engineers should consider during building, training, and testing algorithms (Batarseh, Freeman, and Huang, 2021).

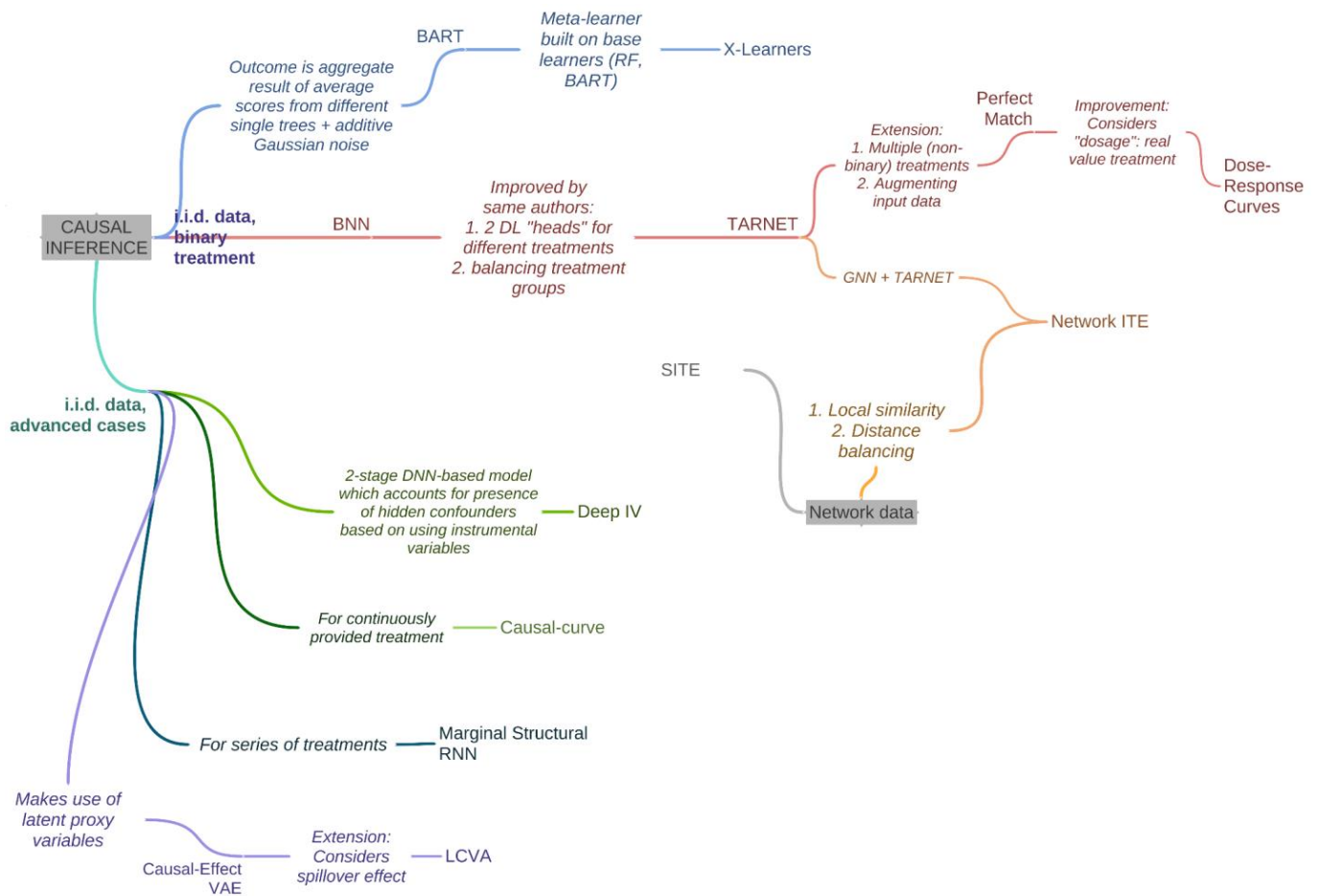

*Figure 3. Literature review knowledge graph (except domain-specific part)*

First, the outcomes produced by an AI system should be trustworthy, otherwise, the fundamental idea of replacing certain human-centered operations with automated intelligent systems would be undermined, which would prevent their further development. Second, the decisions made by the system should be fair and ethical with regards to its users, or to be able to detect, avoid, or eliminate any sort of bias in its outputs. We also want the AI system and its processes to be explainable to in turn ensure trustworthiness, but also be secured from potential outside threats. In certain scenarios, some assurance goals are more important than others. For instance, in military applications ethics gives way to safety, but in civil applications such as automated hiring fairness comes first.

Since the main components in the AI pipeline are the model itself and input data/outcomes, they should be the main target of assurance methods. Input data should be checked for completeness and importance for a particular task AI system created for. Other approaches aim at revealing details of internal processes in the algorithm during the training phase. For instance, deep neural network training mechanism is based on updating its connection weights and controlling the changes of weight values is one of the ways of providing explainability. Moreover, the researchers can develop assurance metrics for each of the categories which can be integrated directly into the objective function. This method largely depends on what AI approach is used and how the model is trained.

In the following section we apply the basics of causal inference to the assurance problem.

3.2. Methods for leveraging causality in assurance

In Section 2, we presented some causal inference methods. Although the goal of most of these studies is to perform an accurate counterfactual inference, where developed algorithms are used for prediction, we can look at causal inference for assurance purposes from a different angle.

As we mentioned earlier, there are 3 strong assumptions that need to be made to find ATE based on counterfactuals. The common support property can be utilized by itself to address the questions of fairness and ethics. This property states that there should be no bias in treatment decisions, and if this property holds, we can assert that the assurance issues of fairness and ethics are addressed via ensuring diversity in deciding what treatments are provided for various patients.

A more comprehensive way to conduct assurance analysis is to build a causal graph and detect connections between features. Such analysis can be useful both at the initial stage of developing an AI model and during its utilization. In the first case, a causal graph can serve as a validation tool for a less explainable AI model and make an educated decision of what features to use as inputs and outputs for this model. This method is in a way similar to imposing constraints on data/model based on real world knowledge, such as physics-guided architecture of neural networks, as proposed by Daw et al. (2020). Section 3.3 describes an experiment performed to leverage this idea.

In the second scenario, the information about causal relationships in the dataset can be analyzed together with the model outputs to provide insights on the meaning of the outcomes given input data. For instance, a parallel metamodel can be run on the main model's outputs to perform calculations of metrics "on the fly" for dynamic analysis, such as mean squared errors, average output values etc.

In Section 3.3, we demonstrate a causal experiment inspired by the first approach, where the feature-focused causal model is created to define inputs and outputs for the future model.

3.3.   Application of Causality in Assurance: Economy of Technology Example

The goal of the experiment is to perform causal analysis to understand how issued tech policies affect the target metric. DoWhy Library (Sharma and Kiciman, 2020) was used to drive causation in a technology policy dataset (Anuga et al., 2021). The main sources of information were the Federal Communications Commission (FCC) (https://www.fcc.gov/) and U.S. Census Bureau (https://www.census.gov/) websites which include tech related laws passed in the U.S. on a state level in a particular year, technology metrics such as number of internet users and average internet speeds, and characteristics of each state.

The aggregate numeric economy of technology dataset includes 341 features divided into 2 broad categories: Environmental Descriptors and Technology Metrics. The former represents contextual information about each state, such as population across different years, land/water areas, and funds available. Technology Metrics include various indicators of prevalence of internet and other technology across different age groups and devices. Another dataset includes law texts together with state and year issued. In this study we combined and wrangled those datasets to match our needs of causal analysis. The statistics of the dataset can be found in Table 1.

After the dataset had been wrangled, we had the following variables:
- *State:* the name of the state
- *Year:* the year the law was passed
- *Funds:* the total amount of money spent on tech development (rendered in millions)
- *Population density*
- *Internet users:* the overall number of internet users of ages 3 and above
- *Treatment*: a binary label indicating whether a tech law was issued (True) or not (False)

*Table 1. Dataset Summary Statistics*

| Feature name | Number of non-empty records |
| --- | --- |
| State | 400 |
| Year | 400 |
| Funds | 300 |
| Population density | 400 |
| Internet users | 250 |
| Treatment | 400 |

In this case, (state, year) tuple is a unique record for our causal model. We utilize *Internet users* variable as the outcome, *Treatment* as treatment provided, and *Funds* and *Population density* as covariates. To consider the dimension of time, we also added *Pre users* and *Post users* columns which contain the number of people using the internet in the preceding and following years accordingly. The rationale behind it is the assumption we make regarding the causal structure: the number of users in the past causes the initiation of the law (*Treatment*), which in turn causes the change in number of users in the future. To handle gaps in *Internet users* column, we implemented cubic interpolation to fill in gaps in years 2013 and 2015 and padding to do so in 2017 (the value similar to the one in 2016 just means the absence of change in key metric from 2016 to 2017). The table excerpt is shown in Table 2, where the total number of records is 300

*Table 2. Tech policy dataset*

|    | State   | Year | Funds | Population_density | Treatment | Pre_users | Post_users |
|----|---------|------|-------|--------------------|-----------|-----------|------------|
| 1  | Alabama | 2013 | 24296 | 95.370708          | FALSE     | 3.03E+06  | 2.96E+06   |
| 2  | Alabama | 2014 | 24577 | 95.602082          | TRUE      | 2.87E+06  | 3.21E+06   |
| 3  | Alabama | 2015 | 24546 | 95.810354          | FALSE     | 2.96E+06  | 3.46E+06   |
| 4  | Alabama | 2016 | 26474 | 96.031065          | TRUE      | 3.21E+06  | 3.51E+06   |
| 5  | Alaska  | 2013 | 11288 | 1.291649           | FALSE     | 5.51E+05  | 5.21E+05   |
| 6  | Alaska  | 2014 | 11397 | 1.290274           | FALSE     | 5.26E+05  | 5.33E+05   |
| 7  | Alaska  | 2015 | 13767 | 1.292403           | FALSE     | 5.21E+05  | 5.38E+05   |
| 8  | Alaska  | 2016 | 10602 | 1.299339           | FALSE     | 5.33E+05  | 5.12E+05   |
| 9  | Arizona | 2013 | 27668 | 58.39005           | FALSE     | 4.32E+06  | 4.10E+06   |
| 10 | Arizona | 2014 | 28905 | 59.249681          | FALSE     | 4.03E+06  | 4.47E+06   |
| 11 | Arizona | 2015 | 31182 | 60.123521          | FALSE     | 4.10E+06  | 4.91E+06   |

Next, we initialized the causal model with the structure demonstrated in Figure 4. In addition to our main nodes, we also accounted for potential unobserved confounders (U) and instrumental variables (Z). Similarly to other covariate features (*Funds, Population_density*) U variable affects treatment and outcome variables: *Post_users, Pre_users*, and *Treatment*. To consider a sequence of events in time as mentioned earlier, we modeled the influence of the number of users in the past (*Pre_users*) on *Treatment,* which in turn affects *Post_users*. The Z variable directly affects only *Treatment* in this experiment.

Then we identified the causal effect qualitatively based on provided dependencies, where the assumption of little importance of unobserved confounders was made. The produced estimate has been utilized to obtain the estimated average treatment effect. We chose the propensity score matching method. DoWhy also supports the "refute estimate" method, allowing researchers to compare the results with the placebo treatment. In this experiment, treatment values (TRUE/FALSE) were replaced with random binary values to emulate causal effect after treatment random permutation. The output is as follows:

```
Refute:            Use a Placebo Treatment
Estimated Effect:  828406.606918239
New Effect:        2876.0043238992657
p value:           0.4749805528770472
```

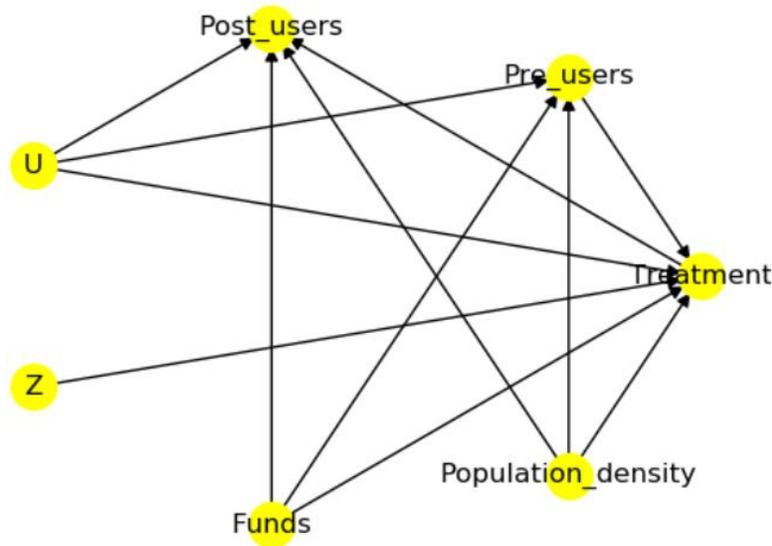

*Figure 4. Causal graph*

As we can see from the numbers, the release of a tech policy leads to an increase of the number of internet users by 828406 on average compared to 2876 in the case of placebo treatment (can be viewed as "idle" legislation). Although the number is not an exact indicator of quantitative effect and varies from one legislation to another, the model is consistent. Based on the results, we can conclude that the number of internet users is a good choice of an output for an AI learning algorithm, whereas other variables can be utilized as inputs given time dimension constraints.

## 4. Network Representations of Data

For many applications it can be beneficial to see the dataset as a network structure through revealing connections between entities. Preserving internal dependencies is one example of how to facilitate in-depth analysis of the dataset and ensure higher transparency during data-related AI stages. We provide an overview of the basic provisions of graph theory and recurrent graph neural networks (RGNN), and finally provide several graph representations of U.S. states and corresponding technology laws from the dataset used in Section 3.

4.1. An Introduction to Graph Theory

In the following section, essential concepts of graph theory are embodied within the concepts of Graph Neural Networks (GNN) which are crucial for building and training a ML algorithm that handles graph data most effectively (F. Scarselli et al., 2009; T. Kipf and M. Welling, 2016; M. Defferrard et al., 2017; W. Hamilton et al., 2017). The core input data structure for a GNN to work is the graph. As alluded to earlier, graphs are formally defined as a set of vertices V along with the set of edges E between these vertices. In standard fashion we define a graph as $G = (V, E)$, where $|V| = N$ is the number of nodes in the graph and $E = N_E$ is the number of edges. We define $A \in R^{NxN}$ as the adjacency matrix related to G (T. Kipf and M. Welling, 2016; M. Defferrard et al., 2017). Fundamentally, graphs are just a way to encode data visually where properties of graphs represent real elements and concepts within the data. Developing insight into how graphs are used as representations of complex concepts is critical in their efficacy as encoding mechanisms or reasoning over features derived from their structure (W. Hamilton et al., 2017).

*Vertices:* In a graph, the objects that are connected are called vertices. These can usually represent entities which are typically defined as attributes with their relationships and how they are connected to other objects. Given a set of *N* vertices denoted as *V,* the $i^{th}$ single vertex we defined as $v_i$(W. Hamilton et al., 2017).

*Edges:* Vertices are connected to one another along edges that characterize the relationships that exist between these vertices. In a strict sense, we defined a single edge between two (not necessarily unique) vertices. Note that a set of $N_E$ edges is denoted as *E* and a single edge between the $i^{th}$ and $j^{th}$ vertices is denoted as $e_{i,j}$(F. Scarselli et al., 2009; T. Kipf and M. Welling, 2016; M. Defferrard et al., 2017; W. Hamilton et al., 2017).

*Features:* In AI, phenomena under study are relegated to quantifiable attributes known as features. Within graph theory for AI, we can utilize these features to express the interactions more deeply between various vertices and edges. In the example of a social network, people are connected to other people, locations, or activities where features for each person (vertex) could quantify the attributes of a person (e.g., age, popularity, and social media usage) (D. Gosnell & M. Broechele, 2020; I. Robinson et al., 2015; M. Needham & A. Hodler, 2019). Furthermore, features that express relationships between vertices (i.e., edges) could include the quantification of the strength of a relationship or affinity (e.g., familial, colleague, etc.). From a feature standpoint, there can be many considerations per vertex and edge; hence, we represent these as vectors expressed as $v_i^F$ and $e_{i,j}^F$ respectively (F. Scarselli et al., 2009; T. Kipf and M. Welling, 2016).

*Neighborhoods:* Neighborhoods are smaller portions of a graph made up of nodes and vertices, defined formally as subgraphs, that can be treated as quite distinct sets of vertices and edges. A neighborhood can be iteratively formed through a single vertex by inspecting all connected vertices and edges connected to it. As a neighborhood grows from the ith vertex $v_i$ it will be denoted as the set of neighbor indices $ne[v_i]$. Note that specific criteria can also be defined by specified criteria for the vertex and edge features (T. Kipf and M. Welling, 2016; M. Defferrard et al., 2017; W. Hamilton et al., 2017; D. Gosnell & M. Broecheler, 2020).

*States*: States are encoded via the information within a given vertices' neighborhood inclusive of the features and states of the neighborhood's vertex and edge). States are defined as "hidden feature vectors" (F. Scarselli et al., 2009). In graph theory these states are iteratively created through a process of extracting features from the previous state's iteration, where classification, regression, or other computation are performed on these iteration states (T. Kipf and M. Welling, 2016; M. Defferrard et al., 2017).

*Embeddings:* Embeddings are representations acquired through reduction of large feature vectors (F. Scarselli et al., 2009). The associated vertices and edges within low dimensional embeddings make it possible to classify them with linearly separable models. The quality of an embedding is measured through the similarity retained in the embedding. Furthermore, these can be "learned" for different parts of the graph (e.g., vertices, edges, neighborhoods, or graphs). Finally, embeddings are also known as representations, encodings, latent vectors, or high-level feature vectors (T. Kipf and M. Welling, 2016; M. Defferrard et al., 2017; W. Hamilton et al., 2017).

4.2     Recurrent Graph Neural Networks (RGNN)

In a standard neural network, successive layers of learned weights work to extract features from an input. After being processed by sequential layers, the resultant high-level features can then be provided to a softmax layer or single neuron for the purpose of classification, regression, etc. A softmax function is often the final neural network activation function that normalizes output of predicted output class probability functions, based on Luce's choice axiom (R. Luce, 1959). Luce's choice axiom addresses "independence from irrelevant alternatives" (IIA) where the selection of an item over another in a pool of many items is not affected by the existence or non-existence of other items in the pool (I. Goodfellow et al., 2016). In this same way, the earliest GNN works aimed to extract high level feature representations from graphs by using successive feature extraction operations (F. Scarselli et al., 2009), and then fed these high-level features to output functions. The recursive application of a feature extractor, or encoding network, is what provides the RGNN with its name (T. Kipf and M. Welling, 2016; M. Defferrard et al., 2017; W. Hamilton et al., 2017).

*The Forward Pass:* The RGNN forward pass occurs in two main steps. The first step focuses on computing high level hidden feature vectors for each vertex in the input graph. This computation is performed by a transition function, $f$. The second step is concerned with processing the hidden feature vectors into useful outputs; using an output function $g$ (T. Kipf and M. Welling, 2016).

*Transition:* The transition process considers the neighborhood of each vertex $v_i$ in a graph, and produces a hidden representation for each of these neighborhoods. Since different vertices in the graph might have different numbers of neighbors, the transition process employs a summation over the neighbors, thus producing a consistently sized vector for each neighborhood. This hidden representation is often referred to as the vertex's state (F. Scarselli et al., 2009), and it is calculated based on the following quantities (M. Defferrard et al., 2017; W. Hamilton et al., 2017; D. Gosnell & M. Broecheler, 2020)

(1) $v_i^F$ - the features of the vertex $v_i$, which the neighborhood is centered around.

(2) $e_{i,j}^F$ - the features of the edges which join $v_i$ to its neighbor vertices $v_j$. Here only direct neighbors are considered, though in practice neighbors further than one edge away may be used. Similarly, for directed graphs, neighbors may or may not be considered based on edge direction (e.g., only outgoing or incoming edges considered as valid neighbor connection).

(3) $v_j^F$ - the features of $v_i$'s neighbors.

(4) $h_j^{k-1}$ - the previous state of $v_i$'s neighbors. Recall that a state simply encodes the information represented in each neighborhood. Formally, the transition function $f$ is used in the recursive calculation of a vertex's $k^{th}$ state as per the following equation:

$$h_i^k = \sum_{j \in ne[v_i]} f(v_i^F, e_{i,j}^F, v_j^F, h_j^{k-1}), \text{ where all } h_i^0 \text{ are defined upon initialization}$$

We can see that under this formulation, $f$ is well defined. It accepts four feature vectors which all have a defined length, regardless of which vertex in the graph is being considered, regardless of the iteration. This means that the transition function can be applied recursively, until a stable state is reached for all vertices in the input graph. If $f$ is a contraction map, Banach's fixed point theorem ensures that the values of $h_i^k$ will converge to stable values exponentially fast, regardless of the initialization of $h_i^0$ (M. Khamsi and W. Kirk, 2001). The iterative passing of messages or states between neighbors to generate an encoding of the graph is what gives this message passing operation its name. In the first iteration, any vertex's state encodes the features of the neighborhood within a single edge. In the second iteration, any vertex's state is an encoding of the features of the neighborhood within two edges away, and so on. This is because the calculation of the $k^{th}$ state relies on the $(k-1)^{th}$ state. To fully elucidate this process, we step through how the transition function is recursively applied. The purpose of repeated applications of the transition function is thus to create discriminative embeddings which can ultimately be used for downstream machine learning tasks.

*Output:* The output function is responsible for taking the converged hidden state of a graph $G(V,E)$ and creating a useful and relevant output. Note that, the transition function $f$ application to features of $G(V,E)$ ensure all final states $h_i^{k_{max}}$ are encoded in some part of $G(V,E)$. The region size dependency centers around the halting condition (convergence, max time steps, etc.), but often the repeated 'message passing' ensures that each vertex's final hidden state has 'seen' the entire graph (F. Scarselli et al., 2009; T. Kipf and M. Welling, 2016). These rich encodings typically have lower dimensionality than the graph's input features and can be fed to fully connected layers for the purpose of the ML technique The output function $g$, akin to $f$ the transition function, is implemented by a feedforward neural network (F. Scarselli, 2009), though other means of returning a single value have been used, including mean operations, dummy super nodes, and attention sums (J. Zhou, 2018; T. Kipf and M. Welling, 2016; M. Defferrard et al., 2017; W. Hamilton et al., 2017). A loss function makes this possible, defined as the error taken from the predicted output and a labelled ground truth (W. Hamilton et al., 2017). Both $f$ and $g$ can then be trained via backpropagation of errors (F. Scarselli et al., 2009) for cases that are relevant.

4.3   Economy of Technology Dataset as a Network

The dataset provides unique structural properties when described as a graph. We shall illustrate three different graph views of the dataset we utilized in Section 3. We shall illustrate graphs of states with respect to title, category, and topics. Figure 5 illustrates a non-directed graph of Categories and States. Figure 6 presents a graph of states and topics in the data.

Observe that in Figure 5 and Figure 6 the density of connection is far more pronounced than in Figure 7. The graph metric that describes how many edges a vertex has is called centrality which helps to determine what vertices could be the most important. Also notice some of the vertices only have one edge which also represents less important vertices. From the standpoint of our dataset, for Figure 5 and Figure 6 that describe topics and categories per state, a graph representation can provide insight into what legal categories and topics are most relevant per state. Note that in formulating a GNN to support any type of correlation, anomaly detection, or prediction, such understanding of how to engineer features is critical in formulating an approach toward any type of AI.

## 5. Conclusion

In this chapter, the authors attempted to familiarize the reader with the concepts of causality and its role in AI assurance, as well as how to build and handle network datasets. We discussed in detail the significance of assuring AI systems applied to the lawmaking process and covered theoretical foundations of causal inference. These concepts were connected through demonstration and explanation of the outcomes of a causal experiment with the economy of technology dataset. We also introduced graph theory, presented examples of structuring the dataset as a network, and elucidated the benefits of such representation. We provided examples of how graph expressions of a sample dataset can provide unique structural insights into the data set. We hope our work inspired the reader to view their AI-applicable and assurance problems from a new angle and supplied them with helpful background and toolset.

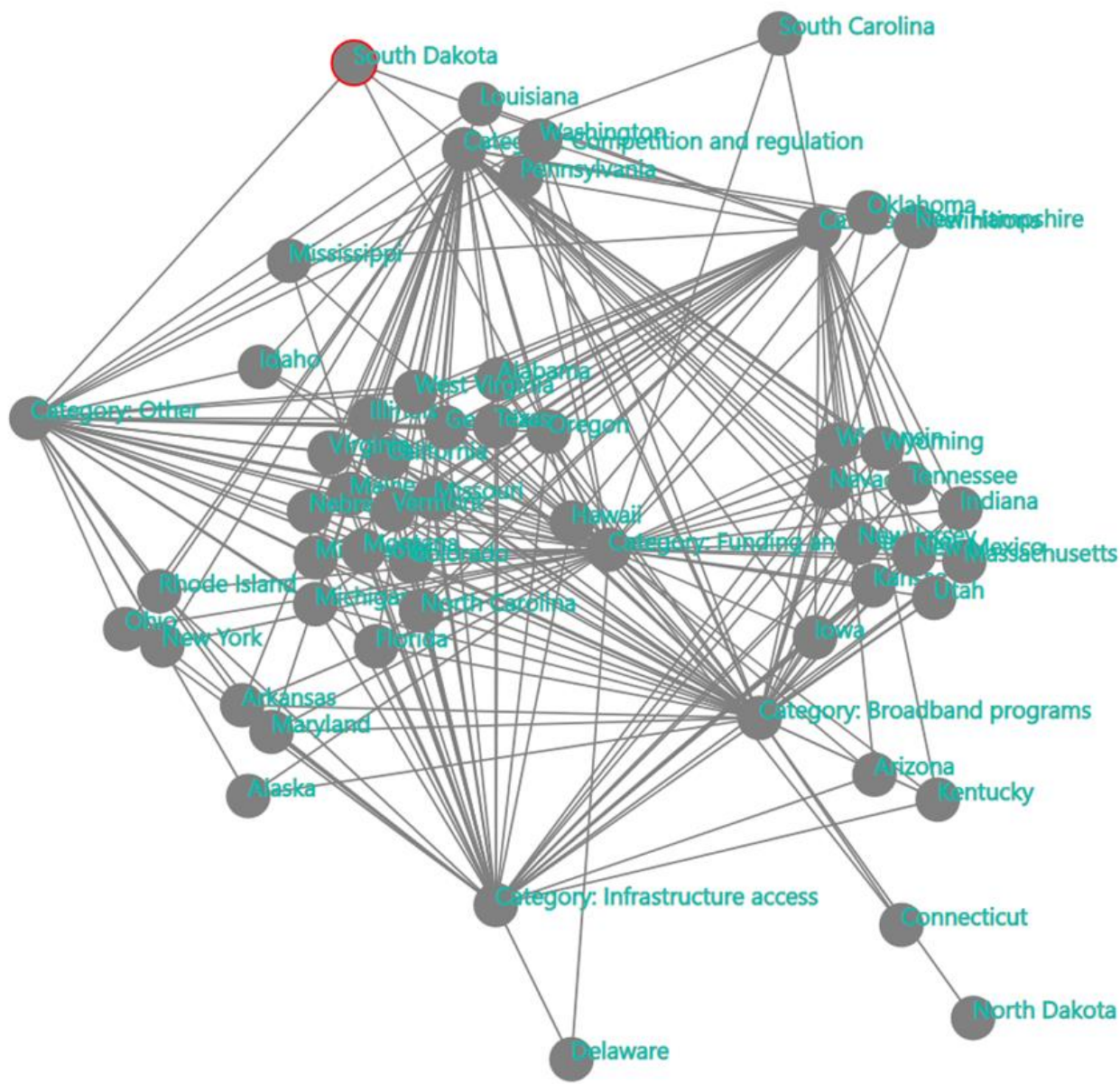

*Figure 5. Graph of Categories and States in the economy of technology dataset*

*Figure 6. Graph representation of states and topics in the economy of technology dataset*

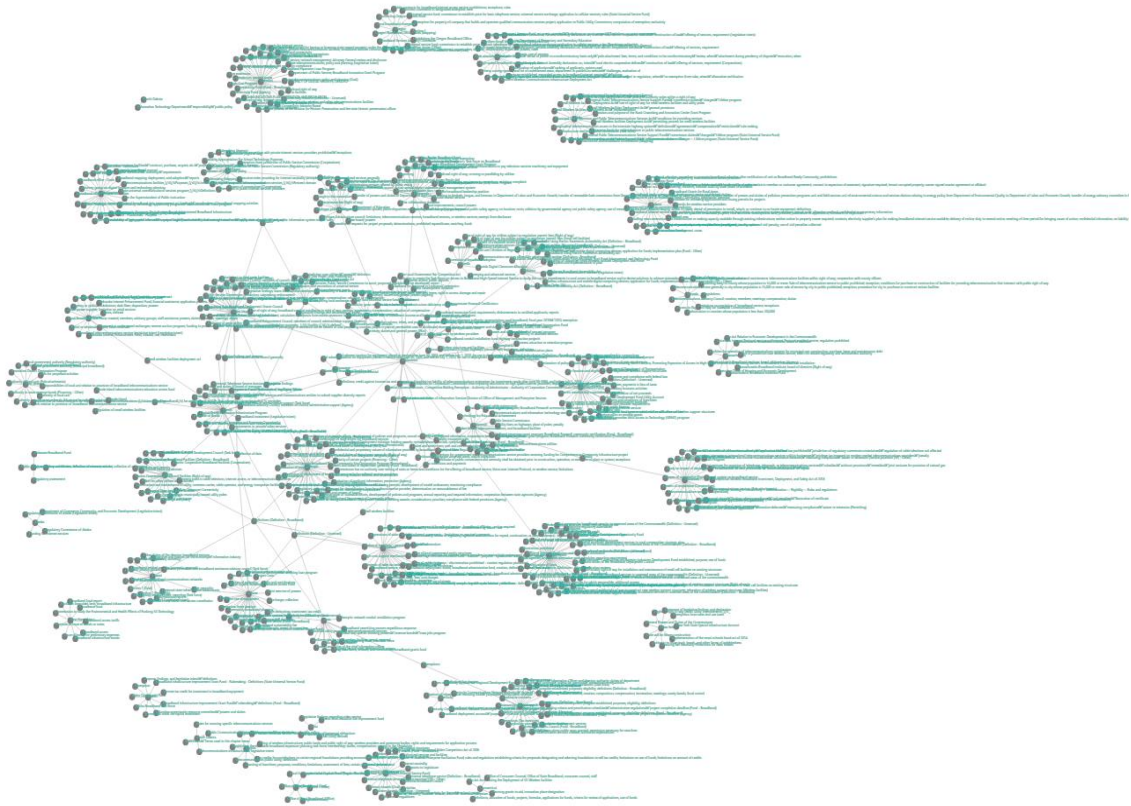

*Figure 7. Graph of U.S. states and titles. Each U.S. state is centered around the related topics that are connected to it. Each cluster represents the U.S. States' respective topics where lines between clusters identify the relationships that exist.*

**Acknowledgements**
The authors would like to thank Dr. Laura Freeman for providing valuable comments on chapter contents and acknowledge Dominick Perini and Amanda Tolman, AI researchers at the A3 research lab (Virginia Tech), for collection and fusion of data used in experimental part of Section 3 of this chapter.